\documentclass{article}

\usepackage{arxiv}

\usepackage[utf8]{inputenc} % allow utf-8 input
\usepackage[T1]{fontenc}    % use 8-bit T1 fonts
\usepackage{hyperref}       % hyperlinks
\usepackage{url}            % simple URL typesetting
\usepackage{booktabs}       % professional-quality tables
\usepackage{amsfonts}       % blackboard math symbols
\usepackage{nicefrac}       % compact symbols for 1/2, etc.
\usepackage{microtype}      % microtypography
\usepackage{lipsum}
\usepackage{graphicx}
\graphicspath{ {./images/} }

\usepackage{amssymb}
\usepackage{amsthm}
\usepackage{amsmath}
\usepackage{amssymb}
\usepackage{makecell}
\usepackage{multirow}

\usepackage{xcolor}
\definecolor{darkgreen}{RGB}{18,178,26}

\newcommand{\RN}[1]{%
	\textup{\uppercase\expandafter{\romannumeral#1}}%
}

\title{SiamixFormer: a fully-transformer Siamese network with temporal Fusion for accurate building detection and change detection in bi-temporal remote sensing images}

\author{
 Amir mohammadian, Foad Ghaderi \\
  Human-Computer interaction lab, Electrical and Computer Engineering Department\\
  Tarbiat Modares University \\
  Tehran, Iran \\
}

\begin{document}
\maketitle
\begin{abstract}
Building detection and change detection using remote sensing images can help urban and rescue planning. Moreover, they can be used for building damage assessment after natural disasters. Currently, most of the existing models for building detection use only one image (pre-disaster image) to detect buildings. This is based on the idea that post-disaster images reduce the model's performance because of presence of destroyed buildings. In this paper, we propose a siamese model, called SiamixFormer, which uses pre- and post-disaster images as input. Our model has two encoders and has a hierarchical transformer architecture. The output of each stage in both encoders is given to a temporal transformer for feature fusion in a way that query is generated from pre-disaster images and (key, value) is generated from post-disaster images. To this end, temporal features are also considered in feature fusion. Another advantage of using temporal transformers in feature fusion is that they can better maintain large receptive fields generated by transformer encoders compared with CNNs. Finally, the output of the temporal transformer is given to a simple MLP decoder at each stage. The SiamixFormer model is evaluated on xBD, and WHU datasets, for building detection and on LEVIR-CD and CDD datasets for change detection and could outperform the state-of-the-art.
\end{abstract}

% keywords can be removed

\section{Introduction}
\label{sec1}
Natural disasters such as earthquakes, floods, and tsunamis have always been a human concern that take lives of many people every year and impose many financial losses on different countries. Highly affected by the changes in global climate, natural disasters have become stronger and more frequent in recent years \cite{rolnick2022tackling}.One of the most important tasks in response phase of disaster management is assessing damages to buildings and roads from satellite images as fast as possible. To this end, it is essential to detect buildings and identify the level of changes in them. Detecting building and their changes can also be used for urban planning, rescue planning, and preparation before disasters. Traditionally, these analyses are performed by experts, which is a time-consuming task \cite{shen2021bdanet}.

With recent advances in computer vision, it is possible to analyze satellite images automatically with high accuracy and speed. In traditional methods, hand-crafted features such as color, shadow, edge, and roof texture were extracted from images, and a feature vector was generated for each sample. Then classification or clustering was done using classic machine learning algorithms \cite{sirmacek2008building, ferraioli2009multichannel, awrangjeb2011improved, gong2013fuzzy}. In more recent techniques, like convolutional neural networks (CNNs), features are extracted in a data-driven manner. Much effort has been done on utilizing CNN models and fusing deep and shallow features, in this problem. For example, a dual-stream network (DS-Net) that adaptively captures local and long-range information is proposed in \cite{zhang2020local}. The authors of \cite{liu2020multiscale} introduced a multi-scale fusion network to tackle the problem of different scales of buildings in remote sensing images. A Siamese fully connected network is proposed in \cite{ji2018fully} and it is shown that using two inputs can help the model obtain better segmentation accuracy. 

Using bi-temporal images is a common approach in problems such as change detection or building damage assessment. In this approach, images that are taken at two different times are compared in order to assess the difference between them. In general, combining the features extracted from bi-temporal images is performed at different levels as follows: 
\begin{enumerate}
	\item 
	\textbf{Segmentation map level}: Two separate encoder-decoder pairs and a differentiating  module are used. Each encoder-decoder pair generates a segmentation map for each bi-temporal image, and the differentiating module is used to compare them and generate the final change map \cite{ji2019building, liu2019temporal}.
	\item 
	\textbf{Feature level}: Two separate encoders, a common decoder and a fusion module are used. The extracted features from each encoder are compared and fused by the fusion module, and the decoder generates the final segmentation map \cite{chen2020spatial,peng2020optical}.
	\item 
	\textbf{Input image level}: Bi-temporal images are combined. An encoder-decoder pair extracts the features from combination of the two images and produces the final change map \cite{de2020change,zhao2020using}. 
\end{enumerate} 

Recently, many researchers from different domains used transformer architectures. These models were first introduced for natural language processing \cite{vaswani2017attention}, however, their application expanded quickly to other domains, e.g., computer vision applications \cite{dosovitskiy2020image, xie2021segformer}. Due to their robust feature presentation, large and global receptive field, and the capability of modeling long-range dependency between pixels, transformer models can be used in remote sensing problems such as building detection and change detection as well \cite{chen2021building}. Xiao \textit{et al.} \cite{xiao2022swin} presented STEB-UNET model, which is a combination of swin transformer and U-Net architectures. Because each building occupies a tiny part of the remote sensing images, Chen \textit{et al.} represented buildings as a set of sparse feature vectors and introduced the sparse token transformer and could reduce computational complexity \cite{chen2021building}. Bandara \textit{et al.} used a hierarchical structure for the transformer encoder and used two encoders to extract features of bi-temporal images \cite{bandara2022transformer}. They concatenated each stage’s output and fed it to an decoder. Chen \textit{et al.} introduced the BIT-CD model by combining CNN and transformer architectures \cite{chen2021remote}.

The authors of \cite{xie2021segformer}, extended the idea of transformers and proposed SegFormer architecture, which is a model for semantic segmentation problems. Although this model has fewer parameters compared with the other transformer-based models, it performs better in benchmark datasets \cite{xie2021segformer}. Considering the success of Segformer, recently, different models based on this architecture have been introduced, e.g., the Damformer \cite{chen2022dual} and the Changeformer \cite{bandara2022transformer} models that have been used for building damage assessment and change detection problems, respectively. In both models, CNNs are used in the feature fusion section.

In this paper, we propose the SiamixFormer model that uses bi-temporal images for building detection. Experimental results confirm that this approach outperforms the existing methods for building and change detection problems. Our proposed model differs from other siamese models in its approach to feature fusion. While both models use a semantic segmentation network for feature extraction, SiamixFormer incorporates a temporal transformer for feature fusion, which takes into account the temporal relationship between features in $T_1$ and $T_2$ images. Unlike other models that use CNNs for feature fusion, this allows Siamixformer to maintain a large receptive field and improve the accuracy of change detection and building detection tasks. The contributions of our work are as follows:
\begin{itemize}
	\item We introduced a novel siamese model, SiamixFormer, that leverages a fully-transformer architecture for semantic segmentation tasks involving bi-temporal input images, such as building detection and change detection from remote sensing data. Our model adopts transformer-based feature extraction and incorporates a temporal transformer for feature fusion, which explicitly accounts for the temporal relationship between features in $T_1$ and $T_2$ images. By doing so, SiamixFormer achieves superior performance in terms of accuracy compared to other state-of-the-art methods, thanks to its ability to maintain a large receptive field and capture both spatial and temporal dependencies in the data.
	\item We utilized bi-temporal images instead of mono-temporal images used by other existing models. Bi-temporal images offer additional information about buildings in the target area over time, enabling our model to capture and analyze pre- and post-disaster images more accurately. The use of bi-temporal images in our model enhances its ability to identify and segment buildings.
	\item Our model is superior to other state-of-the-art methods in terms of F1-score and IoU for building detection and change detection.
\end{itemize}

The structure of the paper is as follows. In section 2, we introduce our proposed model and its details. In section 3, we present the experimental results and compare the performance of our model with that of the state-of-the-art models. Finally, we conclude the paper in section 4.

\begin{figure*}[t]
	\begin{center}
		\includegraphics[width=\linewidth]{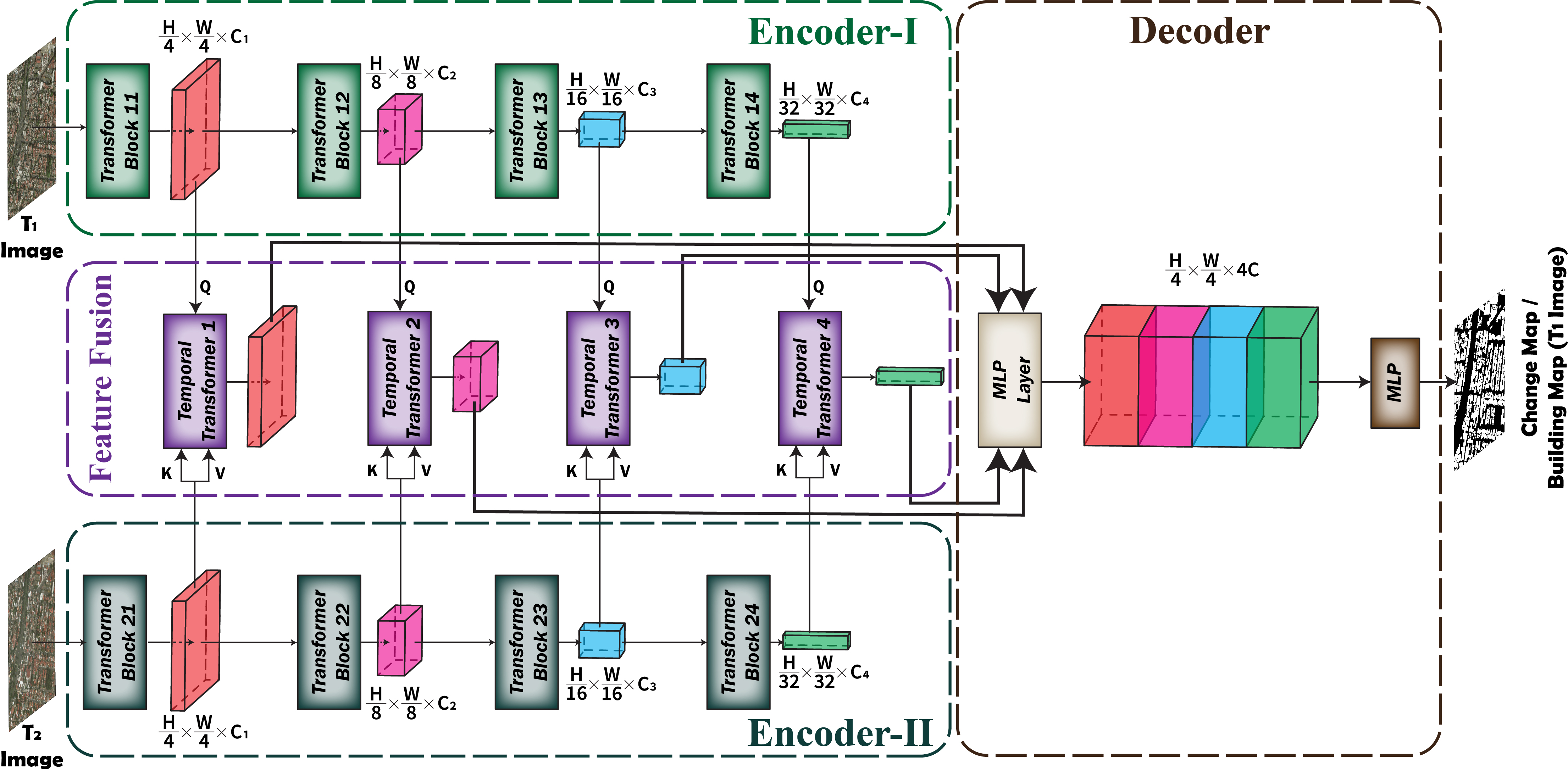}
		\caption{Overview of the proposed SiamixFormer model that consists of four main modules: two encoders that extract features of $T_1$ and $T_2$ input images taken from the same place at two different times; a feature fusion module using a temporal transformer to fuse extracted features considering the temporal feature and their context; and a light-weight decoder. 
			Output of the model for building detection is a segmentation map that indicates the location of buildings in the $T_1$ input image, and the output for change detection is a change map that highlights the differences between the $T_1$ and $T_2$ images.}
		\label{modelfig}
	\end{center}
\end{figure*} 

\section{Methodology}           
\label{sec2}
In this section, we introduce our proposed model, named SiamixFormer, which stands for Siamese mix transformer. First, we explain the data pipeline and the overall architecture of the model. After that, encoder, temporal transformer, and decoder architectures are described. Finally, the loss functions that we used for different datasets are explained.

\subsection{Overall architecture and data pipeline}
As shown in Figure \ref{modelfig}, two input images taken from the same place at two different times are given to the model. For building detection task, the output of the model is a segmentation map corresponding to the $T_1$ image. The output of the model in change detection task is a change map indicating the changes happened between the two images.  
The SiamixFormer model consists of four main modules:
\begin{itemize}
	\item Encoder-\RN{1} dedicated to processing the $T_1$ image.
	\item Encoder-\RN{2} dedicated to processing the $T_2$ image.
	\item Feature fusion module.
	\item Decoder module.
\end{itemize} 
In our model, we use encoders that are consisting of four transformer stages with a hierarchical structure. As depicted in Figure~\ref{modelfig}, height and width of the output of subsequent transformer blocks decrease, while the number of channels increase. The output of the transformer blocks are fed to the feature fusion module as well.

Using bi-temporal images and the two parallel encoders helps us to obtain diverse features from the same place. Taking advantage of the temporal changes between the two input images and their context, the extracted features are fused in the feature fusion module using the temporal transformer. 
Each temporal transformer's output of the feature fusion module is given directly to the decoder, and a segmentation map is generated using a multilayer perceptron (MLP) network. In the building detection problem, the segmentation map highlights the buildings in $T_1$ image, and in the change detection problem, it detects the changes between $T_1$ and $T_2$ images.

\subsection{Encoders}
We used the architecture of SegFormer encoders in our proposed method. The SegFormer architecture is an encoder-decoder model that has a hierarchical transformer structure. Six types of SegFormers exist according to the number of layers in its encoder. SegFormer-B0 is the smallest model for fast inference, and SegFormer-B5 is the largest model for the best performance \cite{xie2021segformer}. In a similar approach, our proposed model has six implementation types, i.e., SiamixFormer-0 to SiamixFormer-5.

The proposed encoder architecture has four transformer blocks and the output of each transformer block has the the dimensionality of $\frac{H}{2^{(i+1)}} \times \frac{W}{2^{(i+1)}} \times C_i$ where \textit{H} and \textit{W} are respectively the height and width of the input image,  $i \in \{1, 2, 3, 4\}$ is the transformer block number, and $C_i$ is number of channel in transformer block-$i$ while $C_{i+1} > C_i$. This provides high-resolution coarse features and low-resolution fine-grained features that help semantic segmentation performance.

In each transformer block, CNN splits the input data into the desired patches with the specified patch size, stride, and padding, which helps the patches to overlap and maintain local continuity around the patches. Then these patches are flattened, and query, key, and value are generated. We use the method suggested in \cite{wang2021pyramid} to reduce computational complexity and generate the new query, key, and value. The formulas used for reducing the length of the sequences are given below:

\begin{equation}
	\begin{gathered}
		X_{new} = MLP(C.R, C)(Reshape(\frac{N}{R}, C .R)(X) )
	\end{gathered}
\end{equation} where \textit{R} is reduction ratio and \textbf{X} denotes the sequence to be reduced, \textit{i.e.}, \textbf{Q}, \textbf{K}, and \textbf{V}. $Reshape(\frac{N}{R}, C .R)(X)$ denotes reshape \textbf{X} to the one with the shape of $\frac{N}{R}\times(C.R)$. $MLP(C_{in}, C_{out})$ denotes a linear layer that takes input with $C_{in}$ channel and generates output with $C_{out}$ channel. So we  have new \textbf{Q}, \textbf{K}, and \textbf{V} with size $(\frac{N}{R}, C)$. This process reduces the computational complexity from $O(N^2)$ to $O(\frac{N^2}{R})$, where $N = H \times W$. In the SiamixFormer, \textit{R} is set to [8,4,2,1] from transformer block-1 to transformer block-4. 

Each transformer block consists of many layers, and each layer consists of some heads. Table~\ref{table0} shows the number of layers and heads for each model in different transformer blocks. The new values of  \textbf{Q}, \textbf{K}, and \textbf{V} are given in parallel to each layer's heads. Heads use the self-attention module, which can be considered as follow:
\begin{equation}
	Attention(Q,K,V) = Softmax(\frac{QK^T}{\sqrt{d_{head}}})V 
\end{equation}
where $d_{head}$ is the head's dimensionality.

The outputs of the parallel heads are concatenated, and multi-head self-attention(MHA) module's output is generated. To consider positional encoding, a 3×3 convolution is used in two layers of MLP, and finally, the output of each layer is generated. Each layer's output is given to the next layer, and this procedure continues until the last layer in transformer block. These steps can be formulated as follows:
\begin{equation}
	\begin{gathered}
		\hat{Z_l}=MHA(Z_{l-1}) + Z_{l-1}, \hspace{20pt} l \in \{1,2, ..., L\} \\
		Z_l = MLP(GELU(Conv_{3\times3}(MLP(\hat{Z_l})))) + \hat{Z_l}, \hspace{20pt}l \in \{1,2, ..., L\} \\
	\end{gathered}
\end{equation}
where $Z_l$ is layer's output, and GELU denotes Gaussian Error Linear Unit activation function \cite{hendrycks2016gaussian}, \textit{L} is number of layer in transformer block, \textit{MHA} is  multi-head self-attention module and MLP is fully connected layer.

\begin{table}[b!]
	\begin{center}
		\caption{Detailed settings of SiamixFormer's encoder. \textit{H}, \textit{L}, and \textit{C} denote the number of heads, the number of layers, and the number of channels in different transformer blocks, respectively.}
		\label{table0}
		\begin{tabular}{c|c|c|c|c}
			\hline
			\hline
			\textbf{model} & \textbf{block-1} & \textbf{block-2} & \textbf{block-3} & \textbf{block-4} \\
			\hline
			
			SiamixFormer-0 &
			\makecell{ $H_1 = 1$\\ $L_1 = 2$ \\ $C_1 = 32$} & 
			\makecell{ $H_2 = 2$\\ $L_2 = 2$ \\ $C_2 = 64$} & 
			\makecell{ $H_3 = 5$\\ $L_3 = 2$ \\ $C_3 = 160$} & 
			\makecell{ $H_4 = 8$\\ $L_4 = 2$ \\ $C_4 = 256$}   \\
			\hline
			
			SiamixFormer-1 &
			\makecell{ $H_1 = 1$\\ $L_1 = 2$ \\ $C_1 = 64$} & 
			\makecell{ $H_2 = 2$\\ $L_2 = 2$ \\ $C_2 = 128$} & 
			\makecell{ $H_3 = 5$\\ $L_3 = 2$ \\ $C_3 = 320$} & 
			\makecell{ $H_4 = 8$\\ $L_4 = 2$ \\ $C_4 = 512$}   \\
			\hline
			
			SiamixFormer-2&
			\makecell{ $H_1 = 1$\\ $L_1 = 3$ \\ $C_1 = 64$} & 
			\makecell{ $H_2 = 2$\\ $L_2 = 3$ \\ $C_2 = 128$} & 
			\makecell{ $H_3 = 5$\\ $L_3 = 6$ \\ $C_3 = 320$} & 
			\makecell{ $H_4 = 8$\\ $L_4 = 3$ \\ $C_4 = 512$}   \\
			\hline
			
			SiamixFormer-3 &
			\makecell{ $H_1 = 1$\\ $L_1 = 3$ \\ $C_1 = 64$} & 
			\makecell{ $H_2 = 2$\\ $L_2 = 3$ \\ $C_2 = 128$} & 
			\makecell{ $H_3 = 5$\\ $L_3 = 18$ \\ $C_3 = 320$} & 
			\makecell{ $H_4 = 8$\\ $L_4 = 3$ \\ $C_4 = 512$}   \\
			\hline
			
			SiamixFormer-4 &
			\makecell{ $H_1 = 1$\\ $L_1 = 3$ \\ $C_1 = 64$} & 
			\makecell{ $H_2 = 2$\\ $L_2 = 8$ \\ $C_2 = 128$} & 
			\makecell{ $H_3 = 5$\\ $L_3 = 27$ \\ $C_3 = 320$} & 
			\makecell{ $H_4 = 8$\\ $L_4 = 3$ \\ $C_4 = 512$}   \\
			\hline
			
			SiamixFormer-5 &
			\makecell{ $H_1 = 1$\\ $L_1 = 3$ \\ $C_1 = 64$} & 
			\makecell{ $H_2 = 2$\\ $L_2 = 6$ \\ $C_2 = 128$} & 
			\makecell{ $H_3 = 5$\\ $L_3 = 40$ \\ $C_3 = 320$} & 
			\makecell{ $H_4 = 8$\\ $L_4 = 3$ \\ $C_4 = 512$}   \\
			\hline
			\hline			
			
		\end{tabular}  
	\end{center}
\end{table}

\subsection{Temporal Transformer}
The temporal transformer is a critical component of the SiamixFormer model that enables accurate building and change detection in bi-temporal remote sensing images, as inspired by \cite{liu2022siamtrans}. The key innovation of this approach lies in its ability to fuse features while considering their temporal relationship, which is essential for bi-temporal tasks. In building detection, the $T_2$ data stream helps the $T_1$ data stream by providing contextual information from the second image, leading to more accurate building detection. In change detection, the temporal transformer acts as a differentiation module that focuses on the differences between the features extracted from $T_1$ and $T_2$ data streams, enabling the identification of changes between the two input images. 

Furthermore, the temporal transformer also preserves the large effective receptive field provided by the transformer encoder, which is crucial for capturing the spatial context of remote sensing images. This allows the model to consider a larger area of the image when making predictions, improving its ability to detect changes and buildings accurately. Additionally, the attention mechanism of transformers is leveraged in the temporal transformer to enable the model to selectively focus on the most informative features from both $T_1$ and $T_2$ data streams, leading to further improvements in bi-temporal tasks. Overall, the temporal transformer is a novel and powerful feature fusion technique that enhances the performance of the SiamixFormer model in bi-temporal remote sensing image analysis. The process of the temporal transformer can be formulated as:
\begin{equation}
	\begin{gathered}
		\hat{Z_{i}} = Attention(Y_{i}^{1}, Y_{i}^{2},Y_{i}^{2}) + Y_{i}^{1}, i \in \{1,2,3,4\}\\
		Z_{i} = MLP(\hat{Z_{i}}) + \hat{Z_{i}}
	\end{gathered}
\end{equation}
where $Y_{i}^{1}$, $Y_{i}^{2}$ denote output of $T_1$ and $T_2$ data stream transformer block-$i$, respectively. Also $Z_{i}$ denotes output of the $i^{th}$ temporal transformer.

\subsection{Decoder}\label{decsec}
Using a hierarchical transformer allows us to have large receptive fields, and hence we can use a lightweight decoder. In our method, first, the output of each temporal transformer in each stage is entered to one MLP module in order to unify the number of channels. Next, the outputs are upsampled to the same size of $\frac{H}{4} \times \frac{W}{4}$, and concatenated to fuse features from different stages. 
Finally, the concatenated feature maps are fed to the last MLP to generate the final output with the size of $\frac{H}{4} \times \frac{W}{4} \times N_{cls}$, where $N_{cls}$ denotes number of the classes. The decoder can be formulated as follows:
\begin{equation}
	\begin{gathered}
		\hat{F_i} = MLP(C_i, C)(Z_{i}), \forall{i} \\ 
		\hat{F_i} = Upsample(\frac{H}{4} \times \frac{W}{4})(\hat{F_i}), \forall{i} \\ 
		F = MLP(4C,C)(Concat(\hat{F_i})), \forall{i} \\
		M = MLP(C,N_{cls})(F)
	\end{gathered}
\end{equation}
where MLP is fully connected layer, and Upsample, Concat denote upsampling and concatenating operators, respectively.

\subsection{Loss function}
SiamixFormer model was utilized for building detection and change detection problems. The datasets we used were imbalanced to varying degrees, resulted in models favoring majority classes in some cases. To overcome this issue, we employed various loss functions and weights for different datasets to ensure that our model learned from both minority and majority classes, resulting in more accurate and balanced predictions. The details are as follows:

\subsubsection{Weighted Cross-entropy}
To explain weighted cross-entropy, assume that the vectorized segmentation map is $\hat{Y}$
which can be represented as:
\begin{equation}
	\hat{Y} = \{\hat{y}_{i}, i = 1, 2,..., H \times W\} , \hspace{20pt} \hat{y}_i \in \{0, 1\}
\end{equation}
where $\hat{y}_i$ represents a pixel in the image. Weighted cross-entropy can be formulated as follows.
\begin{equation}
	L_{WCE} = \frac{1}{H \times W} \sum_{i=1}^{H \times W} w_{cls} . \log(\frac{\exp(\hat{y_i}[cls])}{\sum_{l=0}^{1}\exp(\hat{y_i}[l])})
\end{equation}
where $w_{cls}$ indicates class weights \cite{pihur2007weighted}.
\subsubsection{Focal Loss}
Focal loss is used for datasets that are imbalanced. It applies a modulating term to cross-entropy loss, and the loss value will increase for misclassified data \cite{lin2017focal}. Focal loss can be formulated as follows:
\begin{equation}
	\begin{gathered}
		L_{Focal}(p_{t}) = -(1 - p_{t})^{\gamma} . \log(p_{t}) \\
		p_t= \begin{cases}
			p & \text{if } y=1\\
			1-p & \text{otherwise}
		\end{cases}
	\end{gathered}
\end{equation}
where $y \in \{0,1\}$ specifies the ground-truth class and $p \in [0,1]$ is the model’s estimated probability. $\gamma$ $\geq$ 0 is a hyperparameter and selecting bigger $\gamma$ values, yields to reduced relative loss for well-classified samples.
\subsubsection{Dice Loss}
Dice coefficient is a metric widely used to calculate the similarity between two images. In segmentation problems, it can be adjusted as a loss function and formulated as follows: \cite{sudre2017generalised}:
\begin{equation}
	L_{Dice} = 1 - \frac{2 . Y . softmax(\hat{Y})}{Y + softmax(\hat{Y})}
\end{equation}

We tested several loss functions and weights that were customized for each dataset. In our experiments, we found that the sum of dice and focal loss functions were the most effective for the LEVIR-CD and WHU datasets. For the CDD dataset, we employed a weighted cross-entropy loss, while for the xBD dataset, we used a standard cross-entropy loss. By selecting the optimal loss function for each dataset, we were able to improve the overall performance of our SiamixFormer model. It is worth noting that we evaluated multiple loss functions and only present the most successful ones here.

\section{Experimental Results}           
\label{sec3}
\subsection{Datasets and Evaluation}
In order to investigate the effectiveness of our proposed model on building detection and change detection problems, we conducted the experiments on two different datasets for each problem. The datasets are introduced in the sequel:
\subsubsection{xBD}
xBD is the largest publicly available dataset for building segmentation and disaster damage assessment. In this dataset, satellite images of 19 types of disasters such as earthquakes, floods, wildfires, and hurricanes with the size of 1024$\times$1024 and a resolution of 0.8 m/pixel are collected. The dataset also includes more than 850,000 buildings with an area of more than 450,000 $km^2$ annotated. There are 18,336/1,866/1,866 images and 632,228/109,724/108,784 buildings in this dataset for train, validation, and test, respectively \cite{gupta2019creating}.
\subsubsection{WHU}
The WHU dataset includes aerial images from 2012 and 2016 of areas where a 6.3-magnitude earthquake occurred in 2011. This dataset is labeled for building detection and change detection. It contains 1260 pairs of images for train and 690 pairs of images for test in the size of 512$\times$512. There are also 12,796 buildings in 2012 and 16,077 buildings in 2016 in an area of 20.5 $km^2$ in this dataset. We have used this dataset for building detection by predicting the buildings in 2016 images \cite{ji2018fully} and Due to the GPU memory capacity limitation, we split the images to 256$\times$256 without overlap.

\begin{figure*}[t!]
	\begin{center}
		\includegraphics[width=0.9\linewidth]{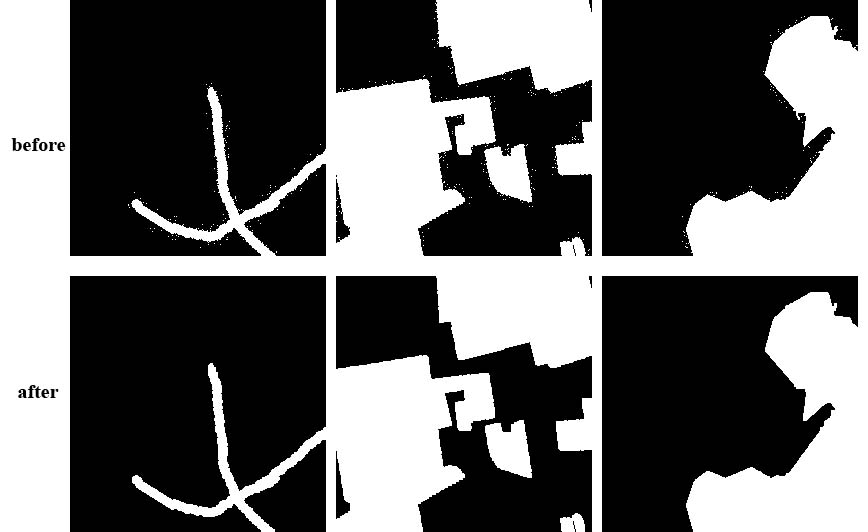}
		\caption{Labels of the CDD dataset, before and after pre-processing.}
		\label{compare_label}
	\end{center}
\end{figure*}

\subsubsection{LEVIR-CD}
This dataset contains 637 pairs of images in the size of $1024 \times 1024$ and a resolution of 0.5 m/pixel, collected from 20 different regions. It focuses on building-related changes, including building growth and building decline. LEVIR-CD covers various buildings such as villa residences, tall apartments, small garages, and large warehouses. This dataset contains 31,333 building changes, with an average of approximately 50 building changes per image and 987 pixels per image change. Due to the GPU memory capacity limitation, we split the images to 256$\times$256 without overlap, as suggested in \cite{chen2021remote}. Finally, we obtained 7120 images for train, 1024 images for validation, and 2048 images for test \cite{chen2020spatial}.

\subsubsection{CDD}
This dataset contains 11 pairs of satellite images with a resolution of 0.03 m/pixel to 1 m/pixel, collected in different seasons. Seven images have the size of 4,725$\times$2,200, and four images with the size of 1,900$\times$1,000, which have been clipped with the size of 256$\times$256. Therefore, we obtain 10,000 images for train, 3000 for validation, and 3000 test images \cite{lebedev2018change}. In this dataset, labels are images in jpg format, and their values are in the range of [0,255] instead of \{0,1\}. To handle this issue, we must consider a threshold and cluster the values into two classes, 0 and 1. By choosing any value for the threshold, some noise is produced in the label images, which makes the learning process challenging. We first considered all values greater than 0 as class 1, which creates much noise; then, we removed these noises by using erosion and dilation. Figure~\ref{compare_label} shows some examples of these images before and after this pre-processing.

\subsubsection{Evaluation metrics}
The experimental results of building detection are evaluated using the F1-score ($F1_{b}$) and intersection over union ($IoU_{b}$) for only building class, which are defined as:
\begin{equation}
	\begin{gathered}
		F1_{b} =\frac{2TP}{2TP + FP + FN}\\
		IoU_{b} =\frac{TP}{TP + FP + FN}\\
	\end{gathered}
\end{equation}
where TP, FP, and FN are the number of true-positive, false-positive, and false-negative pixel of segmentation result, respectively. Moreover, the experimental results of change detection are assessed using the mean of F1-score and IoU for all classes.

\begin{figure*}[b!]
	\begin{center}
		\includegraphics[width=\linewidth]{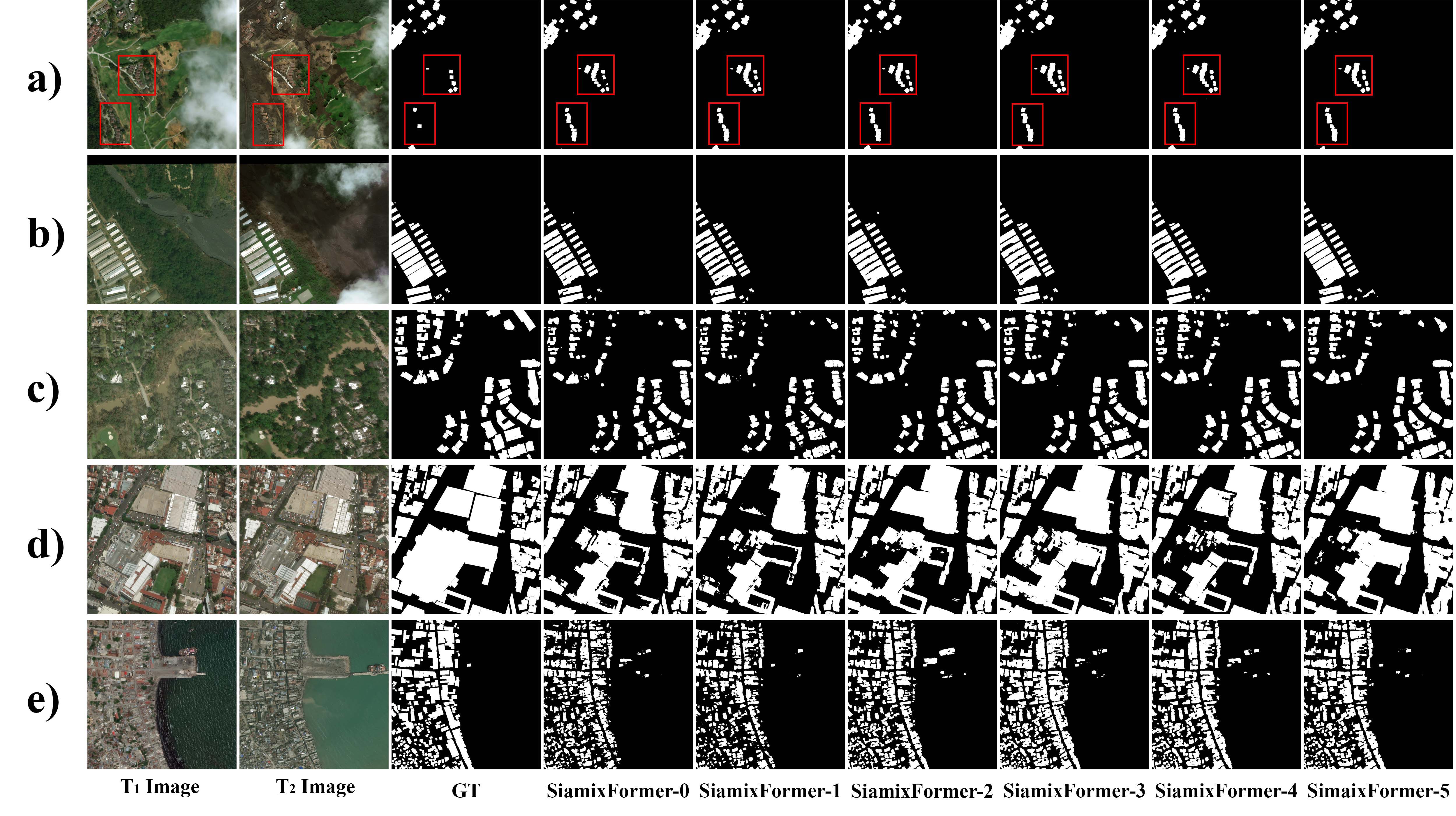}
		\caption{Performance of different SiamixFormer models on the xBD dataset in building detection. In a) red rectangles show some buildings that even though were not labeled in the ground truth (GT) image, the SiamixFormer models succeeded to detect them correctly.}
		\label{compare_xbd}
	\end{center}
\end{figure*}

\subsection{Implementation Details}
We used MMSegmentation \cite{mmseg2020} codebase and trained the models on a single NVIDIA Geforce 1080Ti GPU. For the encoders, we used the pre-trained weights of the SegFormer trained on the ImageNet-1k dataset. The temporal transformer and decoder were randomly initialized. During training, we applied data augmentation through random resize with a ratio of 0.5-2.0 and random horizontal flipping. We also used random cropping with 512$\times$512 for the xBD dataset. The model was trained using an AdamW optimizer for 1M iteration. Due to memory capacity limitations, we used a batch size of 1 for all datasets. Learning rate was set to an initial value of $6\times 10^{-5}$ and a poly LR schedule with the default factor of 1.0 was used.
\begin{table}[b]
	\begin{center}
		\caption{Performance comparison of different method in building detection problem on xBD datasets.}
		\label{table1}
		\begin{tabular}{c|ccccc}
			
			\hline
			\hline
			Method & Input shape & Params(M) & FLOPs(G) & $F1_{b}$ & $IoU_{b}$ \\
			\hline
			RAPNet  & 512$\times$512 & - & - &- &73.26  \\
			BDANet & 512$\times$512 & 34.4& 155.4&86.40 & -  \\
			DamFormer  & 512$\times$512 & - & - &86.86 & - \\
			
			\hline
			SiamixFormer-0 & 512$\times$512 & \textbf{10.04} & \textbf{10.53} & 86.46&76.43 \\
			SiamixFormer-1 & 512$\times$512 & 38.57& 27.71& 87.23&77.35  \\
			SiamixFormer-2 & 512$\times$512 & 60.65&  40.23& 88.05&78.66 \\
			SiamixFormer-3 & 512$\times$512 & 100.41&  63.34& 88.30&79.06 \\
			SiamixFormer-4 & 512$\times$512 & 133.95& 85.59& 88.35& 79.14  \\
			SiamixFormer-5 & 512$\times$512 & 175.15&  108.03& \textbf{88.43}& \textbf{79.26} \\
			\hline
			\hline			
			
		\end{tabular}  
	\end{center}
\end{table}

\begin{figure*}[t!]
	\begin{center}
		\includegraphics[width=\linewidth]{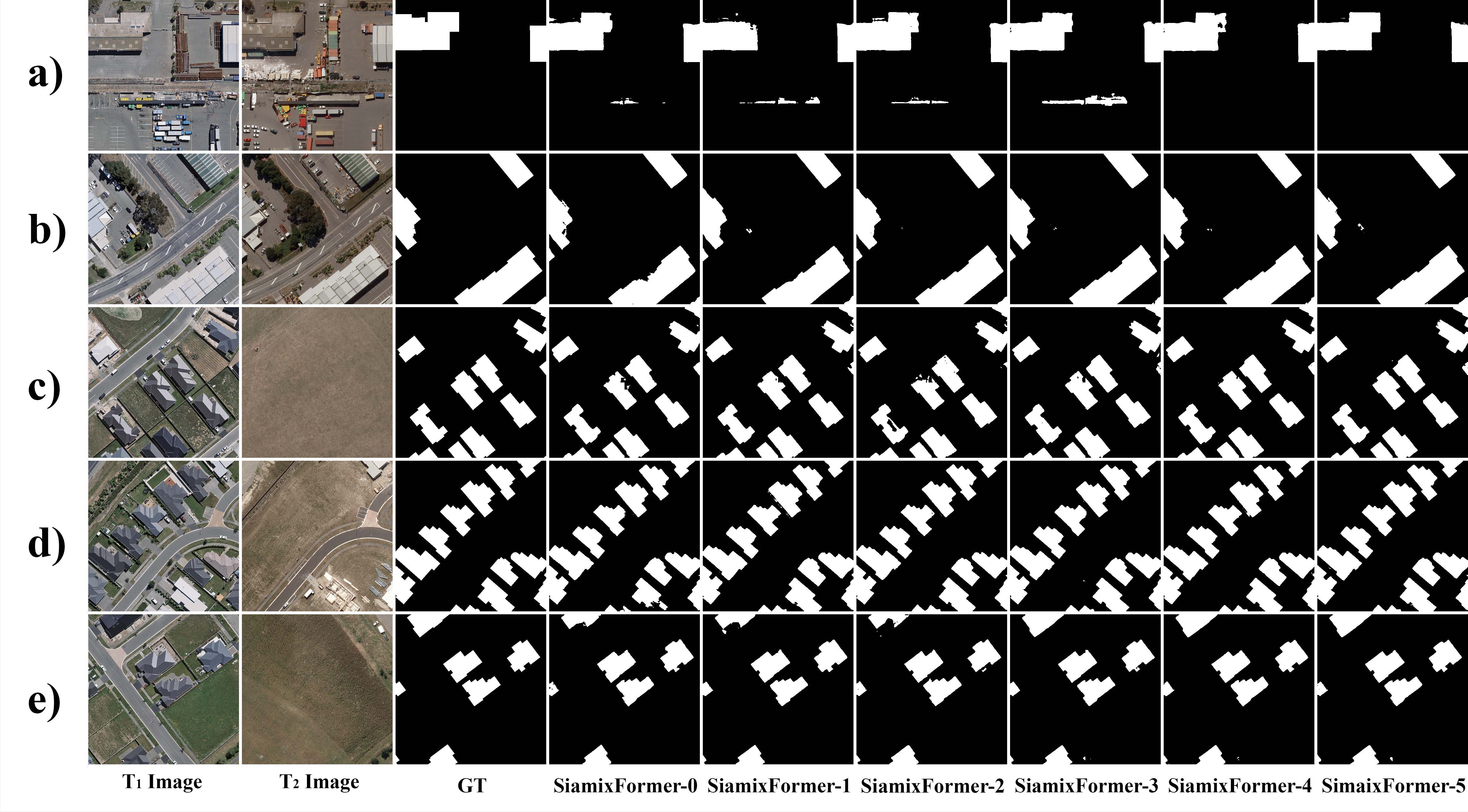}
		\caption{Performance of different SiamixFormer models on the WHU dataset in building detection problem.}
		\label{compare_whu}
	\end{center}
\end{figure*}

\begin{table}[t!]
	\begin{center}
		\caption{Performance comparison of different method in building detection problem on  WHU datasets.}
		\label{table2}
		\begin{tabular}{c|ccccc}
			
			\hline
			\hline
			Method & Input shape & Params(M) & FLOPs(G) & $F1_{b}$ & $IoU_{b}$ \\
			\hline
			DTCDSCN & 256$\times$256 & -& -&- & 88.86\\
			FSAU-Net & 256$\times$256 & -& -&- & 91.73\\
			
			\hline
			SiamixFormer-0 & 256$\times$256 & \textbf{10.04} & \textbf{2.62} & 95.53 &91.43 \\
			SiamixFormer-1 & 256$\times$256 & 38.57&6.91 & 96.01& 92.31 \\
			SiamixFormer-2 & 256$\times$256 & 60.65& 10.06& 96.22& 92.70\\
			SiamixFormer-3 & 256$\times$256 & 100.41& 15.82& 96.32& 92.90\\
			SiamixFormer-4 & 256$\times$256 & 133.95&21.4 & 96.59&93.40 \\
			SiamixFormer-5 & 256$\times$256 & 175.15& 27.01&  \textbf{96.69}&\textbf{93.58} \\
			\hline
			\hline			
			
		\end{tabular}  
	\end{center}
\end{table}

\subsection{Comparison and Analysis}
In this section, we compare the performance of our SiamixFormer model in building detection and change detection with other existing deep learning (CNN-based or transformer-based) models.

\subsubsection{Building Detection}
We considered the following models for building detection:
\begin{itemize}
	\item \textbf{BDANet} \cite{shen2021bdanet}: introduced for building damage assessment, uses a UNet-based model and only pre-disaster images for building detection.
	\item \textbf{RAPNet} \cite{tian2021multiscale}: uses a combination of atrous convolution (AC), deformable convolution (DC), pyramid pooling module (PPM), FPN, and attention mechanism (AM) for building detection.                                                                                                                                                        
	\item \textbf{DamFormer} \cite{chen2022dual}: is based on the SegFormer architecture. It uses CBAM \cite{woo2018cbam} for feature fusion and both pre-disaster and post-disaster images for building detection and building damage assessment.
	\item \textbf{DTCDSCN} \cite{liu2020building}: proposed for change detection. It consists of three parts: two parts perform semantic segmentation, and one part performs change detection. This model has been used for building detection using two inputs on the WHU dataset.
	\item \textbf{FSAU-Net} \cite{hu2023fsau}: proposes a features self-attention U-block network (FSAU-Net) that focuses on the target feature self-attention in the coding stage and introduces spatial attention in the decoder stage to highlight building information areas.
\end{itemize}

Unlike BDANet and RAPNet models, our SiamixFormer model uses two inputs for building detection. As shown in Table \ref{table1}, using pre-disaster and post-disaster images can improve the model's performance in terms of $F1_b$ and $IoU_b$ metrics. This is because the number of destroyed and major-damaged buildings in post-disaster images is small compared to no-damaged and minor-damaged buildings, so using post-disaster images can help detect buildings better. On the other hand, using two inputs can help train the models with large receptive fields better. This way, we can maintain the large receptive field and properly fuse the extracted features. Table \ref{table1} shows that the SiamixFormer model using the temporal transformer in the feature fusion section obtained better results compared with the DamFormer model, which uses CBAM for feature fusion. In Figure \ref{compare_xbd}, the qualitative results of different SimixFormer models on the holdout images of the xBD dataset are presented.

\begin{figure*}[b!]
	\begin{center}
		\includegraphics[width=\linewidth]{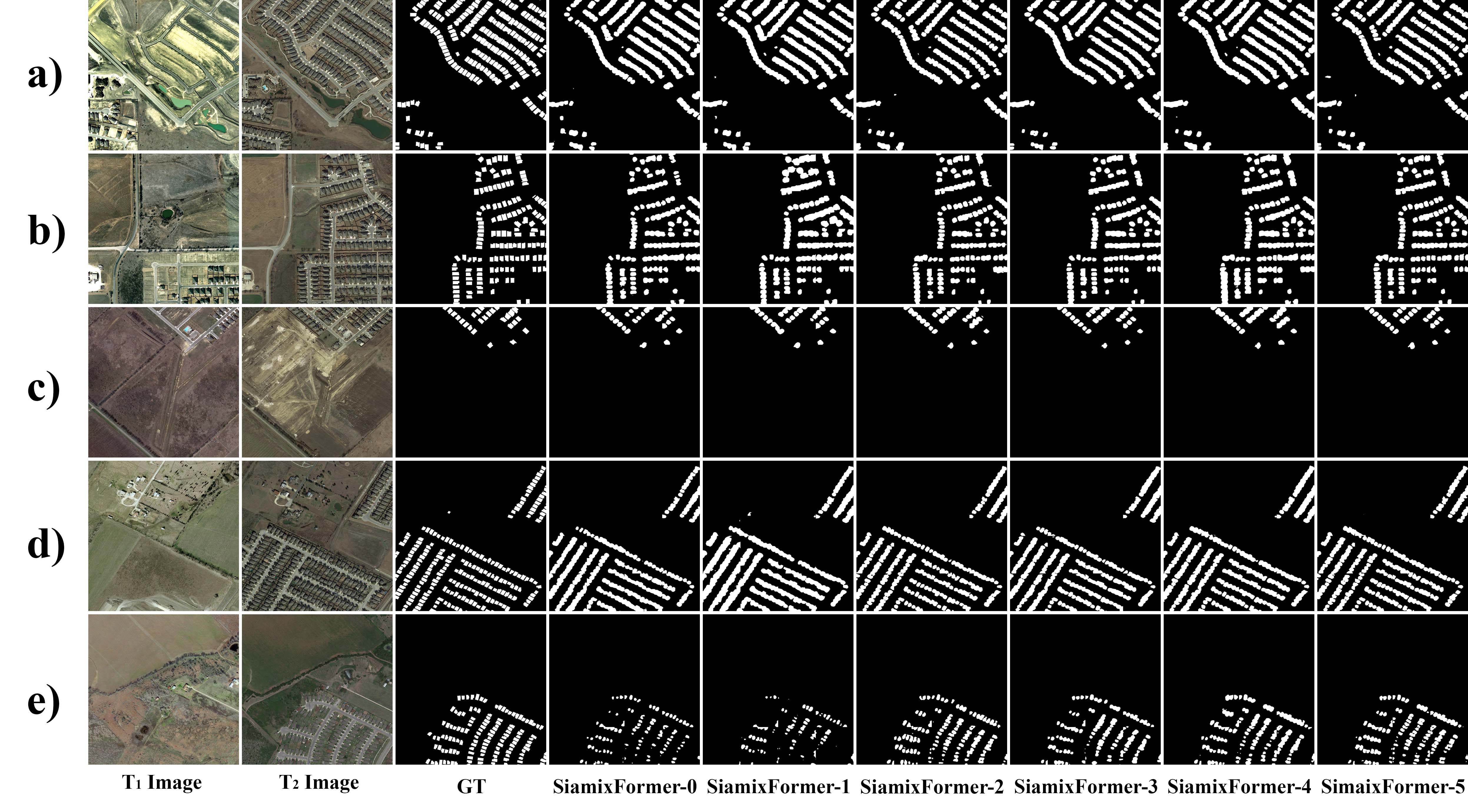}
		\caption{Performance of different SiamixFormer models on the Levir-CD dataset in change detection problem.}
		\label{compare_levir}
	\end{center}
\end{figure*}

\begin{table}[b]
	\begin{center}
		\caption{Performance comparison in change detection task on LEVIR-CD dataset.}
		\label{table3}
		\begin{tabular}{c|ccccc}
			\hline
			\hline
			
			Method & Input shape & Params(M) & FLOPs(G) & $F1$ & $IoU$ \\
			\hline
			UVACD  & 256$\times$256& -& -& 91.30 & 83.98 \\
			FDOR-Net  & 256$\times$256& -& -& 90.85 & 91.13 \\
			FTN  & 256$\times$256& - & 45& 91.01 & 83.51 \\
			SNUNet & 256$\times$256 & 27.06 & 250 &88.16 &78.83 \\
			BIT  & 256$\times$256 & \textbf{3.55}&4.35 &89.31 &80.68  \\
			BIT(IMP-ViTAEv2-S) & 256$\times$256& -& -& 91.26&-\\
			ChangeFormer  & 256$\times$256& 29.75 & 21.19& 90.40 &82.48 \\
			\hline
			SiamixFormer-0 &256$\times$256 & 10.04 &  \textbf{2.62}  &89.47&82.29  \\
			SiamixFormer-1 &256$\times$256 & 38.57&6.91 &   89.29&82.03  \\
			SiamixFormer-2 &256$\times$256 & 60.65& 10.06&   90.57&83.88 \\
			SiamixFormer-3 &256$\times$256 & 100.41& 15.82&   90.54&83.84 \\
			SiamixFormer-4 &256$\times$256 & 133.95&21.4 &   90.70 & 84.05   \\
			SiamixFormer-5 &256$\times$256 & 175.15& 27.01&    \textbf{91.58} & \textbf{85.38}\\
			\hline
			\hline			
		\end{tabular}  
	\end{center}
\end{table}

\begin{table}[t]
	\begin{center}
		\caption{Performance comparison in change detection task and CDD dataset.}
		\label{table4}
		\begin{tabular}{c|ccccc}
			\hline
			\hline
			Method & Input shape & Params(M) & FLOPs(G) & $F1$ & $IoU$ \\
			\hline
			DSAMNET &256$\times$256 & -&- &93.69 & 88.13 \\
			SNUNet & 256$\times$256& 27.06& 250 &96.2 & - \\
			BIT  & 256$\times$256 & \textbf{3.55}&4.35 &90.73 &83.03  \\
			BIT(IMP-ViTAEv2-S) &256$\times$256 &- &- &97.02&-\\
			ChangeFormer  & 256$\times$256& 29.75 & 21.19& 89.38 &80.79 \\
			\hline
			SiamixFormer-0 &256$\times$256 &10.04 &  \textbf{2.62} &92.05&85.81 \\
			SiamixFormer-1 &256$\times$256 & 38.57&6.91  &92.78& 87.00 \\
			SiamixFormer-2 &256$\times$256 & 60.65& 10.06 &95.52& 91.62\\
			SiamixFormer-3 &256$\times$256 & 100.41& 15.82 &96.48&93.33\\
			SiamixFormer-4 &256$\times$256 & 133.95&21.4  &96.85 &94.00 \\
			SiamixFormer-5 &256$\times$256 & 175.15& 27.01 &\textbf{97.13}& \textbf{94.51}\\
			\hline
			\hline			
		\end{tabular}  
	\end{center}
\end{table}

\begin{figure*}[t!]
	\begin{center}
		\includegraphics[width=\linewidth]{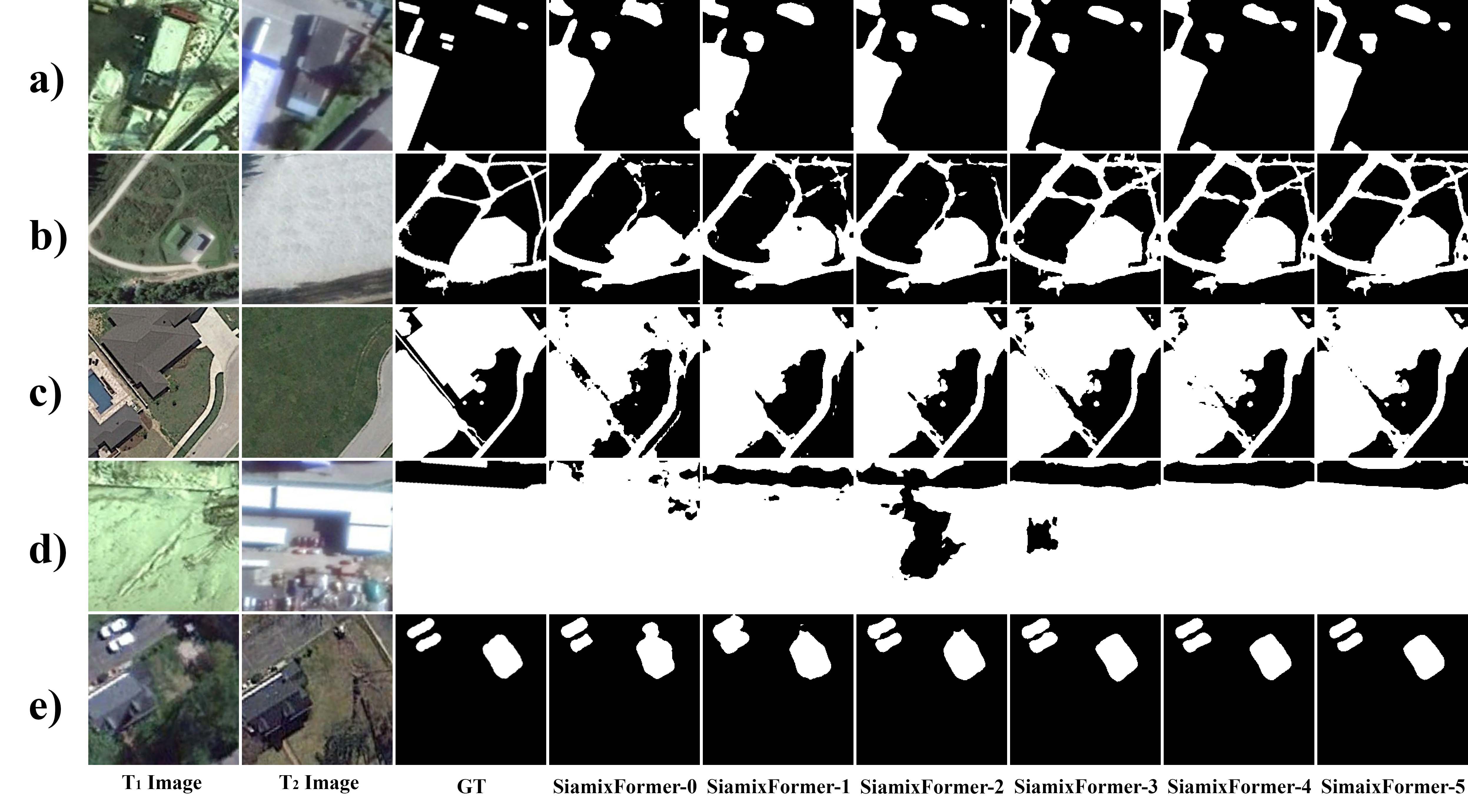}
		\caption{Performance of different SiamixFormer models on the CCD dataset in change detection problem.}
		\label{compare_cdd}
	\end{center}
\end{figure*}

In order to show the robustness of our proposed model in building detection, we also evaluate model on the WHU dataset. This dataset is usually used for change detection, but the DTCDSCN and FSAU-Net models used this dataset for building detection and change detection. As can be seen in Table \ref{table2}, the SiamixFormer model has also achieved outstanding results on this dataset. Figure \ref{compare_whu} shows the qualitative results of different SimixFormer models on some sample images from the test set of the WHU dataset.

\subsubsection{Change Detection}
We considered the following models to compare with our proposed model in change detection problem:
\begin{itemize}
	\item \textbf{DSAMNET} \cite{shi2021deeply}: fuses different levels of extracted features using metric module and CBAM.
	\item \textbf{SNUNet} \cite{fang2021snunet}: based on NestedUNet, fuses the extracted features using the ensemble channel attention module.
	\item \textbf{BIT} \cite{chen2021remote}: uses the CNN backbone and converts the extracted features into semantic tokens. Then, passed the tokens to transformer encoder and generates a change map with transformer decoder. Feature differencing module is the last stage.
	\item \textbf{BIT(IMP-ViTAEv2-S)} \cite{wang2022empirical} : used BIT structure with ViTAEv2-S backbone.
	\item \textbf{ChangeFormer} \cite{bandara2022transformer}: is based on SegFormer and concatenates the features extracted from each stage for feature fusion.
	\item \textbf{UVACD} \cite{wang2022network}: fuses the features extracted from the CNN backbone using a visual transformer.
	\item \textbf{FDOR-Net} \cite{ye2022feature}: The method is based on a Feature Decomposition-Optimization-Reorganization Network (FDOR-Net), which decomposes the input image into different feature maps and optimizes them for building change detection. The network then reorganizes the feature maps to generate a change detection map.
	\item \textbf{FTN} \cite{yan2022fully}: The paper proposes a method that utilizes the Swin transformer network as an encoder and a pyramid structure grafted with a Progressive Attention Module (PAM) as a CNN based feature fusion module.
\end{itemize}

\begin{figure*}[b!]
	\begin{center}
		\includegraphics[width=\linewidth]{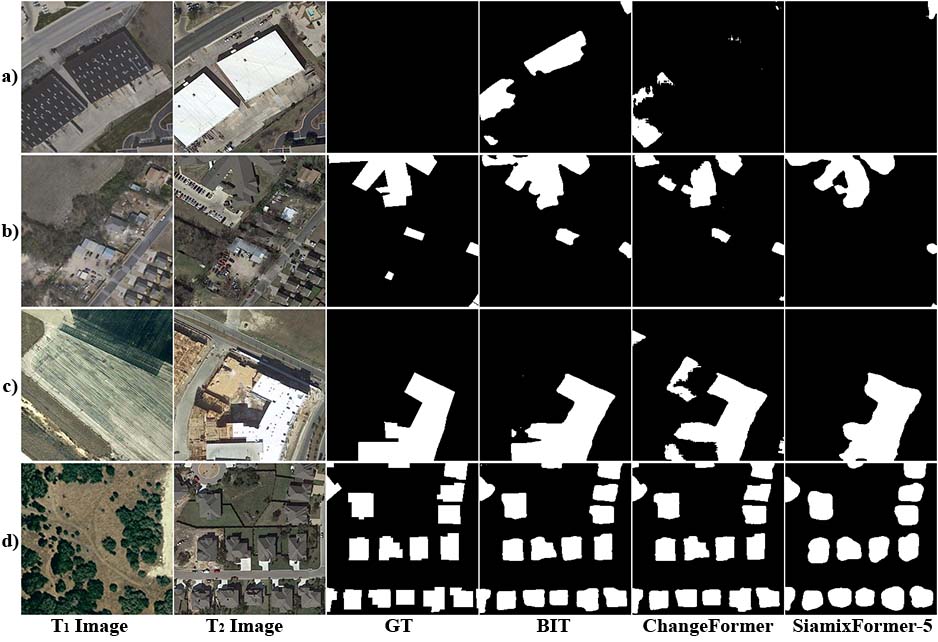}
		\caption{Compare result of SiamixFomer-5 with BIT and ChangeFormer on LEVIR-CD dataset.}
		\label{compare_images}
	\end{center}
\end{figure*}
Most of the models introduced for change detection problem use two decoders to generate a segmentation map for each image and afterwards use another module to detect the changes. These modules analyze the difference between the two segmentation maps. However, we used our SiamixFormer model for change detection without any alteration to the model that was designed for building detection. As shown in Table \ref{table3} and \ref{table4}, the SiamixFormer model achieved promising results on both LEVIR-CD and CDD datasets compared with the other methods. Figures \ref{compare_levir} and \ref{compare_cdd} show the qualitative results of different SimixFormer models on some samples of the test sets of these datasets. To compare the qualitative results of our model with other models, we compared the results of SiamixFormer-5 with models for which model weights are available, such as BIT and ChangeFormer. The comparative results on the LEVIR-CD dataset are shown in the Figure \ref{compare_images}.

A notable point about the SiamixFormer model is that the temporal transformer, which is used for feature fusion in the model, fuses the features in building detection tasks in a way that the $T_2$ stream outputs assist the $T_1$ stream in correctly detecting buildings. On the other hand, in change detection tasks, this module fuses features of the two streams in such a way that the difference between them is given to the decoder.

\begin{figure*}[t]
	\begin{center}
		\includegraphics[width=\linewidth]{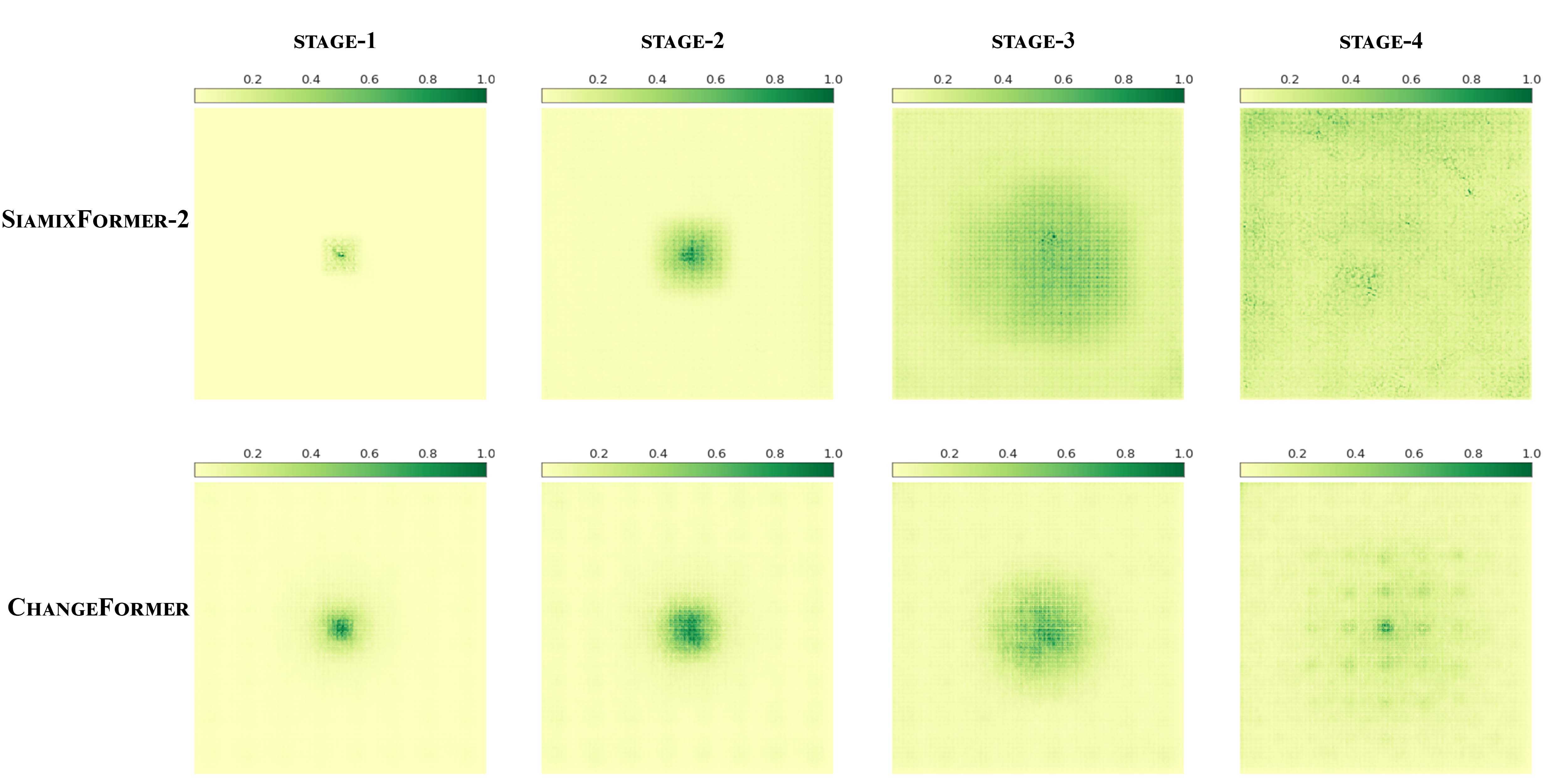}
		\caption{Comparison effective receptive field of each stage's output between SiamixFormer-2 and ChangeFormer.}
		\label{erf}
	\end{center}
\end{figure*}

\subsection{Discussion}
In order to analyze the usefulness of different parts of the proposed architecture, we investigated the effects of using bi-temporal images and using the temporal transformers at different stages of the feature fusion module in experiments conducted on the xBD dataset. As shown in Table \ref{table5}, the SiamixFormer-5 can improve $2.24 \%$ the $F1_b$ metric compared with the SegFormer-B5. This confirms the effect of using pre-disaster and post-disaster images in improved building detection. In another attempt, we used a CNN model to extract features, a temporal transformer (TT) for feature fusion and SegFormer-B5 architecture. The SiamixFormer-5 could achieve $5.61 \%$ better $F1_b$ than this model. This shows that using bi-temporal inputs can help the model if the encoder module can produce features with large receptive fields and the feature fusion module can maintain large receptive fields.

Figure \ref{erf} compares the effective receptive field resulting from the output of each stage between Changeformer and SiamixFormer-2 models on the test set of the LEVIR-CD dataset. Both models have almost used segformer as an encoder, and their difference is in feature fusion. Nevertheless, the SiamixFormer-2 model has preserved a larger effective receptive field than ChangeFormer.

\begin{table}[h!]
	\begin{center}
		\caption{Ablation study on the effects of using pre-disaster and post-disaster images and using the temporal transformers (TT) at stages of the feature fusion module. Experiments were conducted on the xBD dataset.} 
		\label{table5}
		\begin{tabular}{c|cc|c}
			\hline
			\hline
			Method & pre-disaster & post-disaster & $F1_{b}$ \\
			\hline
			SegFormer-B5 & \checkmark & $\times$ & 86.19 \\
			CNN + TT + SegFormer-B5 & \checkmark & \checkmark & 82.82 \\
			SiamixFormer-5 & \checkmark & \checkmark & \textbf{88.43} \\
			\hline
			\hline			
		\end{tabular}  
	\end{center}
\end{table}

\section{Conclusions}           
\label{sec4}
In this paper, we proposed a fully-transformer model with a hierarchical architecture named SiamixFormer. The model was used for building detection and change detection on four different datasets. The SiamixFormer model uses two SegFormer-based encoders with four stages, each receiving one of the bi-temporal images and extracting features with large receptive fields. In each stage, the output of the encoders, in addition to being given to the next stage, is also given to the temporal transformer of the same stage. The temporal transformer is used for feature fusion and achieves this target by exploiting temporal features. Moreover, the temporal transformer can maintain large receptive fields. Experimental results show that using bi-temporal images and temporal transformers in feature fusion can improve the model's performance in both building detection and change detection. One limitation of our proposed method is the number of model parameters, particularly in the SiamixFormer-5 architecture, which may require a larger GPU memory. In future work, we plan to focus on optimizing the model from this perspective, while also exploring how the proposed model can be applied to building classification in building damage assessment problems.

\bibliographystyle{unsrt}  
\bibliography{references}  %%% Remove comment to use the external .bib file (using bibtex).

\begin{thebibliography}{10}

\bibitem{rolnick2022tackling}
David Rolnick, Priya~L Donti, Lynn~H Kaack, Kelly Kochanski, Alexandre Lacoste,
  Kris Sankaran, Andrew~Slavin Ross, Nikola Milojevic-Dupont, Natasha Jaques,
  Anna Waldman-Brown, et~al.
\newblock Tackling climate change with machine learning.
\newblock {\em ACM Computing Surveys (CSUR)}, 55(2):1--96, 2022.

\bibitem{shen2021bdanet}
Yu~Shen, Sijie Zhu, Taojiannan Yang, Chen Chen, Delu Pan, Jianyu Chen, Liang
  Xiao, and Qian Du.
\newblock {BDANet}: Multiscale convolutional neural network with
  cross-directional attention for building damage assessment from satellite
  images.
\newblock {\em IEEE Transactions on Geoscience and Remote Sensing}, 60:1--14,
  2021.

\bibitem{sirmacek2008building}
Beril Sirmacek and Cem Unsalan.
\newblock Building detection from aerial images using invariant color features
  and shadow information.
\newblock In {\em 2008 23rd international symposium on computer and information
  sciences}, pages 1--5. IEEE, 2008.

\bibitem{ferraioli2009multichannel}
Giampaolo Ferraioli.
\newblock Multichannel insar building edge detection.
\newblock {\em IEEE Transactions on Geoscience and Remote Sensing},
  48(3):1224--1231, 2009.

\bibitem{awrangjeb2011improved}
Mohammad Awrangjeb, Chunsun Zhang, and Clive~S Fraser.
\newblock Improved building detection using texture information.
\newblock {\em International Archives of Photogrammetry, Remote Sensing and
  Spatial Information Sciences}, 38:143--148, 2011.

\bibitem{gong2013fuzzy}
Maoguo Gong, Linzhi Su, Meng Jia, and Weisheng Chen.
\newblock Fuzzy clustering with a modified {MRF} energy function for change
  detection in synthetic aperture radar images.
\newblock {\em IEEE Transactions on Fuzzy Systems}, 22(1):98--109, 2013.

\bibitem{zhang2020local}
Hongyan Zhang, Yue Liao, Honghai Yang, Guangyi Yang, and Liangpei Zhang.
\newblock A local-global dual-stream network for building extraction from
  very-high-resolution remote sensing images.
\newblock {\em IEEE Transactions on Neural Networks and Learning Systems},
  2020.

\bibitem{liu2020multiscale}
Yuanyuan Liu, Dingyuan Chen, Ailong Ma, Yanfei Zhong, Fang Fang, and Kai Xu.
\newblock Multiscale {U-shaped CNN} building instance extraction framework with
  edge constraint for high-spatial-resolution remote sensing imagery.
\newblock {\em IEEE Transactions on Geoscience and Remote Sensing},
  59(7):6106--6120, 2020.

\bibitem{ji2018fully}
Shunping Ji, Shiqing Wei, and Meng Lu.
\newblock Fully convolutional networks for multisource building extraction from
  an open aerial and satellite imagery data set.
\newblock {\em IEEE Transactions on Geoscience and Remote Sensing},
  57(1):574--586, 2018.

\bibitem{ji2019building}
Shunping Ji, Yanyun Shen, Meng Lu, and Yongjun Zhang.
\newblock Building instance change detection from large-scale aerial images
  using convolutional neural networks and simulated samples.
\newblock {\em Remote Sensing}, 11(11):1343, 2019.

\bibitem{liu2019temporal}
Ruoyun Liu, Monika Kuffer, and Claudio Persello.
\newblock The temporal dynamics of slums employing a {CNN-based} change
  detection approach.
\newblock {\em Remote sensing}, 11(23):2844, 2019.

\bibitem{chen2020spatial}
Hao Chen and Zhenwei Shi.
\newblock A spatial-temporal attention-based method and a new dataset for
  remote sensing image change detection.
\newblock {\em Remote Sensing}, 12(10):1662, 2020.

\bibitem{peng2020optical}
Xueli Peng, Ruofei Zhong, Zhen Li, and Qingyang Li.
\newblock Optical remote sensing image change detection based on attention
  mechanism and image difference.
\newblock {\em IEEE Transactions on Geoscience and Remote Sensing},
  59(9):7296--7307, 2020.

\bibitem{de2020change}
Pablo~Pozzobon De~Bem, Osmar~Ab{\'\i}lio de~Carvalho~Junior, Renato
  Fontes~Guimar{\~a}es, and Roberto~Arnaldo Trancoso~Gomes.
\newblock Change detection of deforestation in the brazilian amazon using
  landsat data and convolutional neural networks.
\newblock {\em Remote Sensing}, 12(6):901, 2020.

\bibitem{zhao2020using}
Wenzhi Zhao, Xi~Chen, Xiaoshan Ge, and Jiage Chen.
\newblock Using adversarial network for multiple change detection in bitemporal
  remote sensing imagery.
\newblock {\em IEEE Geoscience and Remote Sensing Letters}, 2020.

\bibitem{vaswani2017attention}
Ashish Vaswani, Noam Shazeer, Niki Parmar, Jakob Uszkoreit, Llion Jones,
  Aidan~N Gomez, {\L}ukasz Kaiser, and Illia Polosukhin.
\newblock Attention is all you need.
\newblock {\em Advances in neural information processing systems}, 30, 2017.

\bibitem{dosovitskiy2020image}
Alexey Dosovitskiy, Lucas Beyer, Alexander Kolesnikov, Dirk Weissenborn,
  Xiaohua Zhai, Thomas Unterthiner, Mostafa Dehghani, Matthias Minderer, Georg
  Heigold, Sylvain Gelly, et~al.
\newblock An image is worth 16x16 words: Transformers for image recognition at
  scale.
\newblock {\em arXiv preprint arXiv:2010.11929}, 2020.

\bibitem{xie2021segformer}
Enze Xie, Wenhai Wang, Zhiding Yu, Anima Anandkumar, Jose~M Alvarez, and Ping
  Luo.
\newblock Segformer: Simple and efficient design for semantic segmentation with
  transformers.
\newblock {\em Advances in Neural Information Processing Systems}, 34, 2021.

\bibitem{chen2021building}
Keyan Chen, Zhengxia Zou, and Zhenwei Shi.
\newblock Building extraction from remote sensing images with sparse token
  transformers.
\newblock {\em Remote Sensing}, 13(21):4441, 2021.

\bibitem{xiao2022swin}
Xiao Xiao, Wenliang Guo, Rui Chen, Yilong Hui, Jianing Wang, and Hongyu Zhao.
\newblock A swin transformer-based encoding booster integrated in u-shaped
  network for building extraction.
\newblock {\em Remote Sensing}, 14(11):2611, 2022.

\bibitem{bandara2022transformer}
Wele Gedara~Chaminda Bandara and Vishal~M Patel.
\newblock A transformer-based siamese network for change detection.
\newblock {\em arXiv preprint arXiv:2201.01293}, 2022.

\bibitem{chen2021remote}
Hao Chen, Zipeng Qi, and Zhenwei Shi.
\newblock Remote sensing image change detection with transformers.
\newblock {\em IEEE Transactions on Geoscience and Remote Sensing}, 2021.

\bibitem{chen2022dual}
Hongruixuan Chen, Edoardo Nemni, Sofia Vallecorsa, Xi~Li, Chen Wu, and Lars
  Bromley.
\newblock Dual-tasks siamese transformer framework for building damage
  assessment.
\newblock {\em arXiv preprint arXiv:2201.10953}, 2022.

\bibitem{wang2021pyramid}
Wenhai Wang, Enze Xie, Xiang Li, Deng-Ping Fan, Kaitao Song, Ding Liang, Tong
  Lu, Ping Luo, and Ling Shao.
\newblock Pyramid vision transformer: A versatile backbone for dense prediction
  without convolutions.
\newblock In {\em Proceedings of the IEEE/CVF International Conference on
  Computer Vision}, pages 568--578, 2021.

\bibitem{hendrycks2016gaussian}
Dan Hendrycks and Kevin Gimpel.
\newblock Gaussian error linear units (gelus).
\newblock {\em arXiv preprint arXiv:1606.08415}, 2016.

\bibitem{liu2022siamtrans}
Lin Liu, Shanxin Yuan, Jianzhuang Liu, Xin Guo, Youliang Yan, and Qi~Tian.
\newblock Siamtrans: Zero-shot multi-frame image restoration with pre-trained
  siamese transformers.
\newblock In {\em Proceedings of the AAAI Conference on Artificial
  Intelligence}, volume~36, pages 1747--1755, 2022.

\bibitem{pihur2007weighted}
Vasyl Pihur, Susmita Datta, and Somnath Datta.
\newblock Weighted rank aggregation of cluster validation measures: a monte
  carlo cross-entropy approach.
\newblock {\em Bioinformatics}, 23(13):1607--1615, 2007.

\bibitem{lin2017focal}
Tsung-Yi Lin, Priya Goyal, Ross Girshick, Kaiming He, and Piotr Doll{\'a}r.
\newblock Focal loss for dense object detection.
\newblock In {\em Proceedings of the IEEE international conference on computer
  vision}, pages 2980--2988, 2017.

\bibitem{sudre2017generalised}
Carole~H Sudre, Wenqi Li, Tom Vercauteren, Sebastien Ourselin, and
  M~Jorge~Cardoso.
\newblock Generalised dice overlap as a deep learning loss function for highly
  unbalanced segmentations.
\newblock In {\em Deep learning in medical image analysis and multimodal
  learning for clinical decision support}, pages 240--248. Springer, 2017.

\bibitem{gupta2019creating}
Ritwik Gupta, Bryce Goodman, Nirav Patel, Ricky Hosfelt, Sandra Sajeev, Eric
  Heim, Jigar Doshi, Keane Lucas, Howie Choset, and Matthew Gaston.
\newblock Creating xbd: A dataset for assessing building damage from satellite
  imagery.
\newblock In {\em Proceedings of the IEEE/CVF conference on computer vision and
  pattern recognition workshops}, pages 10--17, 2019.

\bibitem{lebedev2018change}
MA~Lebedev, Yu~V Vizilter, OV~Vygolov, VA~Knyaz, and A~Yu Rubis.
\newblock Change detection in remote sensing images using conditional
  adversarial networks.
\newblock {\em International Archives of the Photogrammetry, Remote Sensing \&
  Spatial Information Sciences}, 42(2), 2018.

\bibitem{mmseg2020}
MMSegmentation Contributors.
\newblock {MMSegmentation}: Openmmlab semantic segmentation toolbox and
  benchmark.
\newblock \url{https://github.com/open-mmlab/mmsegmentation}, 2020.

\bibitem{tian2021multiscale}
Qinglin Tian, Yingjun Zhao, Yao Li, Jun Chen, Xuejiao Chen, and Kai Qin.
\newblock Multiscale building extraction with refined attention pyramid
  networks.
\newblock {\em IEEE Geoscience and Remote Sensing Letters}, 19:1--5, 2021.

\bibitem{woo2018cbam}
Sanghyun Woo, Jongchan Park, Joon-Young Lee, and In~So Kweon.
\newblock Cbam: Convolutional block attention module.
\newblock In {\em Proceedings of the European conference on computer vision
  (ECCV)}, pages 3--19, 2018.

\bibitem{liu2020building}
Yi~Liu, Chao Pang, Zongqian Zhan, Xiaomeng Zhang, and Xue Yang.
\newblock Building change detection for remote sensing images using a dual-task
  constrained deep siamese convolutional network model.
\newblock {\em IEEE Geoscience and Remote Sensing Letters}, 18(5):811--815,
  2020.

\bibitem{hu2023fsau}
Minghong Hu, Jiatian Li, Yunfei Zhao, Mei Lu, and Wen Li.
\newblock Fsau-net: a network for extracting buildings from remote sensing
  imagery using feature self-attention.
\newblock {\em International Journal of Remote Sensing}, 44(5):1643--1664,
  2023.

\bibitem{shi2021deeply}
Qian Shi, Mengxi Liu, Shengchen Li, Xiaoping Liu, Fei Wang, and Liangpei Zhang.
\newblock A deeply supervised attention metric-based network and an open aerial
  image dataset for remote sensing change detection.
\newblock {\em IEEE transactions on geoscience and remote sensing}, 60:1--16,
  2021.

\bibitem{fang2021snunet}
Sheng Fang, Kaiyu Li, Jinyuan Shao, and Zhe Li.
\newblock Snunet-cd: A densely connected siamese network for change detection
  of vhr images.
\newblock {\em IEEE Geoscience and Remote Sensing Letters}, 19:1--5, 2021.

\bibitem{wang2022empirical}
Di~Wang, Jing Zhang, Bo~Du, Gui-Song Xia, and Dacheng Tao.
\newblock An empirical study of remote sensing pretraining.
\newblock {\em IEEE Transactions on Geoscience and Remote Sensing}, 2022.

\bibitem{wang2022network}
Guanghui Wang, Bin Li, Tao Zhang, and Shubi Zhang.
\newblock A network combining a transformer and a convolutional neural network
  for remote sensing image change detection.
\newblock {\em Remote Sensing}, 14(9):2228, 2022.

\bibitem{ye2022feature}
Yuanxin Ye, Liang Zhou, Bai Zhu, Chao Yang, Miaomiao Sun, Jianwei Fan, and
  Zhitao Fu.
\newblock Feature decomposition-optimization-reorganization network for
  building change detection in remote sensing images.
\newblock {\em Remote Sensing}, 14(3):722, 2022.

\bibitem{yan2022fully}
Tianyu Yan, Zifu Wan, and Pingping Zhang.
\newblock Fully transformer network for change detection of remote sensing
  images.
\newblock In {\em Proceedings of the Asian Conference on Computer Vision},
  pages 1691--1708, 2022.

\end{thebibliography}

%\begin{thebibliography}{1}
%
%
%\end{thebibliography}

\end{document}